\documentclass[letterpaper]{article} 
\usepackage{aaai24}  
\usepackage{times}  
\usepackage{helvet}  
\usepackage{courier}  
\usepackage[hyphens]{url}  
\usepackage{graphicx} 
\usepackage{natbib}  
\usepackage{caption} 
\usepackage{subfig}
\usepackage{algorithm}
\usepackage{algorithmic}
\usepackage{xcolor}
\usepackage{bbding}
\usepackage{amsmath}
\usepackage{amssymb}
\usepackage{amsthm}
\usepackage{multirow}

\usepackage{newfloat}
\usepackage{listings}
\usepackage{graphicx}
\usepackage{pifont}

\makeatletter

\newcommand{\Rmnum}[1]{\expandafter\@slowromancap\romannumeral #1@}
\makeatother

\DeclareCaptionStyle{ruled}{labelfont=normalfont,labelsep=colon,strut=off} 
\floatstyle{ruled}
\newfloat{listing}{tb}{lst}{}
\floatname{listing}{Listing}
\title{Invariant Representation via Decoupling Style and Spurious Features from Images}
\author {
    Ruimeng Li\textsuperscript{\rm 1,\rm 3},
    Yuanhao Pu\textsuperscript{\rm 1,\rm 3},
    Zhaoyi Li\textsuperscript{\rm 2,\rm 3},
    Hong Xie\textsuperscript{\rm 2,\rm 3, *},
    Defu Lian\textsuperscript{\rm 1,\rm 2,\rm 3,*}
}
\affiliations {
    \textsuperscript{\rm 1}School of Data Science, University of Science and Technology of China\\
    \textsuperscript{\rm 2}School of Computer Science and Technology, University of Science and Technology of China\\
    \textsuperscript{\rm 3}State Key Laboratory of Cognitive Intelligence, Hefei, Anhui, China\\
    lamorak@mail.ustc.edu.cn, puyuanhao@mail.ustc.edu.cn, lizhaoyi777@mail.ustc.edu.cn, xiehong2018@foxmail.com, liandefu@ustc.edu.cn
}

\begin{document}

\maketitle


\begin{abstract}

This paper considers the out-of-distribution (OOD) generalization problem under the setting that both style distribution shift and spurious features exist and domain labels are missing. This setting frequently arises in real-world applications and is underlooked because previous approaches mainly handle either of these two factors. The critical challenge is decoupling style and spurious features in the absence of domain labels. To address this challenge, we first propose a structural causal model (SCM) for the image generation process, which captures both style distribution shift and spurious features. The proposed SCM enables us to design a new framework called IRSS, which can gradually separate style distribution and spurious features from images by introducing adversarial neural networks and multi-environment optimization, thus achieving OOD generalization. Moreover, it does not require additional supervision (e.g., domain labels) other than the images and their corresponding labels. Experiments on benchmark datasets demonstrate that IRSS outperforms traditional OOD methods and solves the problem of Invariant risk minimization (IRM) degradation, enabling the extraction of invariant features under distribution shift. 

\end{abstract}

\section{Introduction}

Statistical correlations, presented in the training dataset, are commonly leveraged by many machine learning approaches to make accurate predictions. These correlations represent subtle patterns and dependencies that play a crucial role in the predictive capability of the models. Learning correlations to make predictions is an effective paradigm under the independent and identically distributed (i.i.d.) hypothesis, which states that the testing and training datasets are independently and identically sampled from the same distribution. However, real-world applications often violate this hypothesis due to the complex data generation mechanisms, including selection biases, confounding factors, or other peculiarities \cite{torralba2011unbiased}. These issues, in practice, are often formulated as the out-of-distribution (OOD) generalization problems.

An increasing number of approaches  \cite{ganin2015unsupervised, arjovsky2019invariant, stablenet} were proposed to address the OOD generalization problem. These approaches follow the principle of adapting or generalizing the model to different domains via learning domain-invariant features. Furthermore, several public benchmark OOD datasets were developed to facilitate the evaluation of OOD generalization approaches, including VLCS \shortcite{fang2013unbiased}, PACS \shortcite{li2017deeper}, OfficeHome \shortcite{venkateswara2017deep}, and NICO \shortcite{he2021towards}.
 
One foundational assumption of previous OOD generalization approaches is that the images within the same domain can be treated as, i.i.d. samples from a fixed but unknown domain-specific distribution. However, the practice of this assumption in real-world applications is hindered by erroneous domain labels or even missing domain labels. For example, definitions of domains across some public datasets need to be more consistent. One reason is that defining domains and partitioning datasets across domains to satisfy the i.i.d. property require sufficient prior domain knowledge and much human effort, which is usually difficult to provide in practice. Facing this challenge, we study the setting that domain labels are missing. 

\begin{figure}[htb]
\centering
\includegraphics[width=.95\columnwidth]{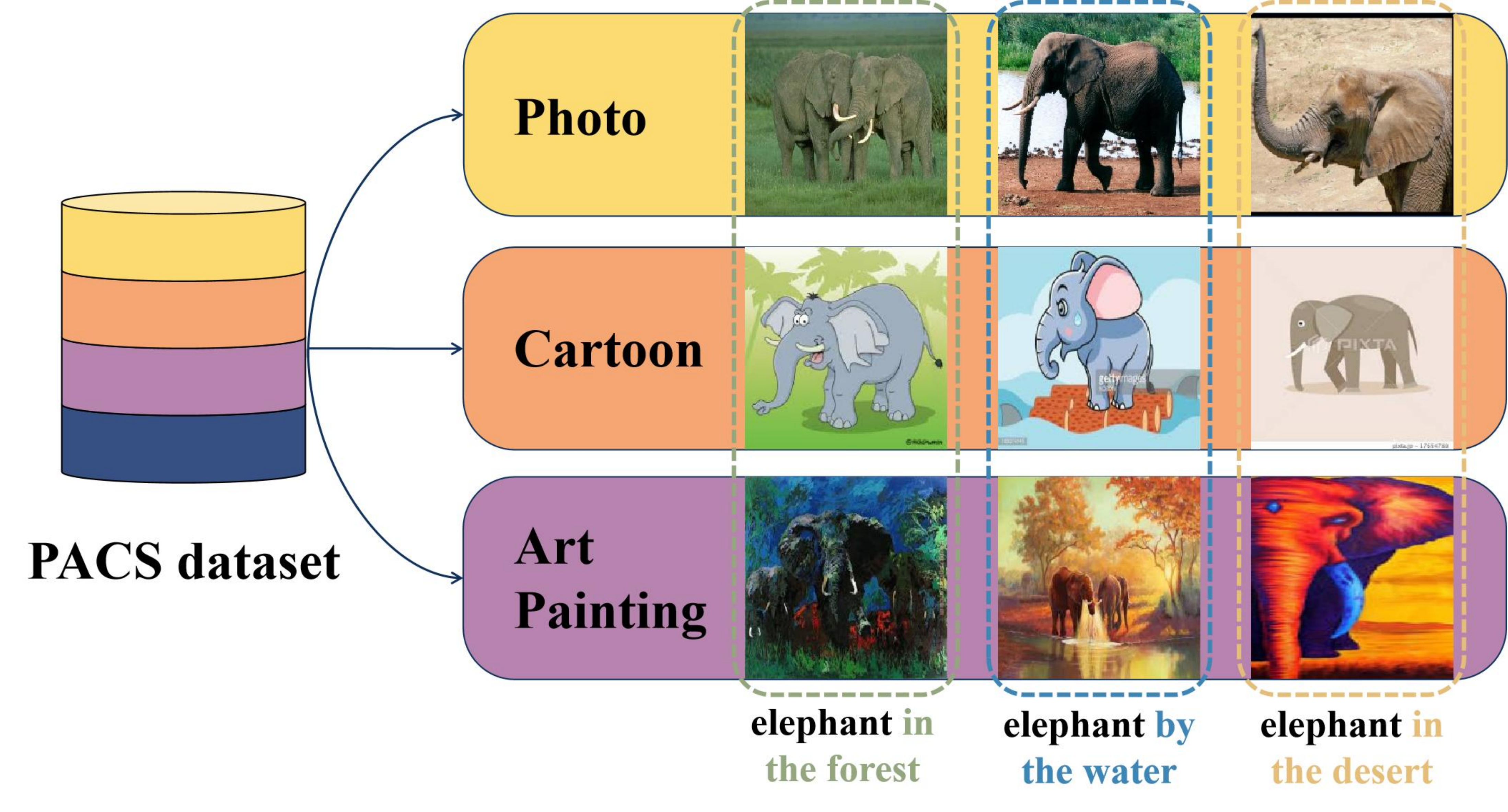} 
\caption{Illustrating OOD problems caused by style and spurious features 
using elephant-labeled images in the PACS dataset \shortcite{li2017deeper}.  
There are two distinct OOD problems: (1) inherent distribution shift in  domain-specific classification due to varying styles across different domains, 
and (2) distribution shift in spurious features across images with the same style, where non-target objects may differ and result in specific distribution shifts.}
\label{oodexample}
\end{figure}

To address the challenge mentioned above, we first analyze several public OOD benchmark datasets to reveal two essential factors that lead to distribution shift: diverse styles and spurious features (i.e., objects outside the target class of interest). Figure \ref{oodexample} illustrates two atomic sub-problems of OOD, which reveal the impact of each factor on the distribution shift individually. We further elaborate on it as follows: 

\begin{itemize}
\item 
{\bf Out-of-distribution problems caused by style.} 
This sub-problem of OOD considers the setting that variations in the style cause the underlying distribution shift. To illustrate, Figure \ref{oodexample} shows that the PACS dataset's inherent domain is divided into four categories, i.e., Photo, Cartoon, Art Painting, and Sketch. In the first image of the Photo and Cartoon categories, despite depicting an elephant in a forest, the style difference leads to a significant distribution shift. This form of distribution discrepancy in OOD is called style distribution shift.

\item {\bf Out-of-distribution problems caused by spurious features.}  
This sub-problem of OOD considers the setting that the underlying distribution shift is caused by spurious features, which refer to objects outside the target class of interest. To illustrate, Figure \ref{oodexample} shows that in the PACS dataset, even within the same style, non-target objects in different images may vary, as exemplified by the first and second images of the Cartoon category. Although both pictures conform to the Cartoon style, the presence of distinct non-target objects leads to a significant distribution shift. This form of distribution discrepancy in OOD is called spurious feature shift.
\end{itemize}

\begin{figure}[htb]
    \centering
    \includegraphics[width=0.5\columnwidth]{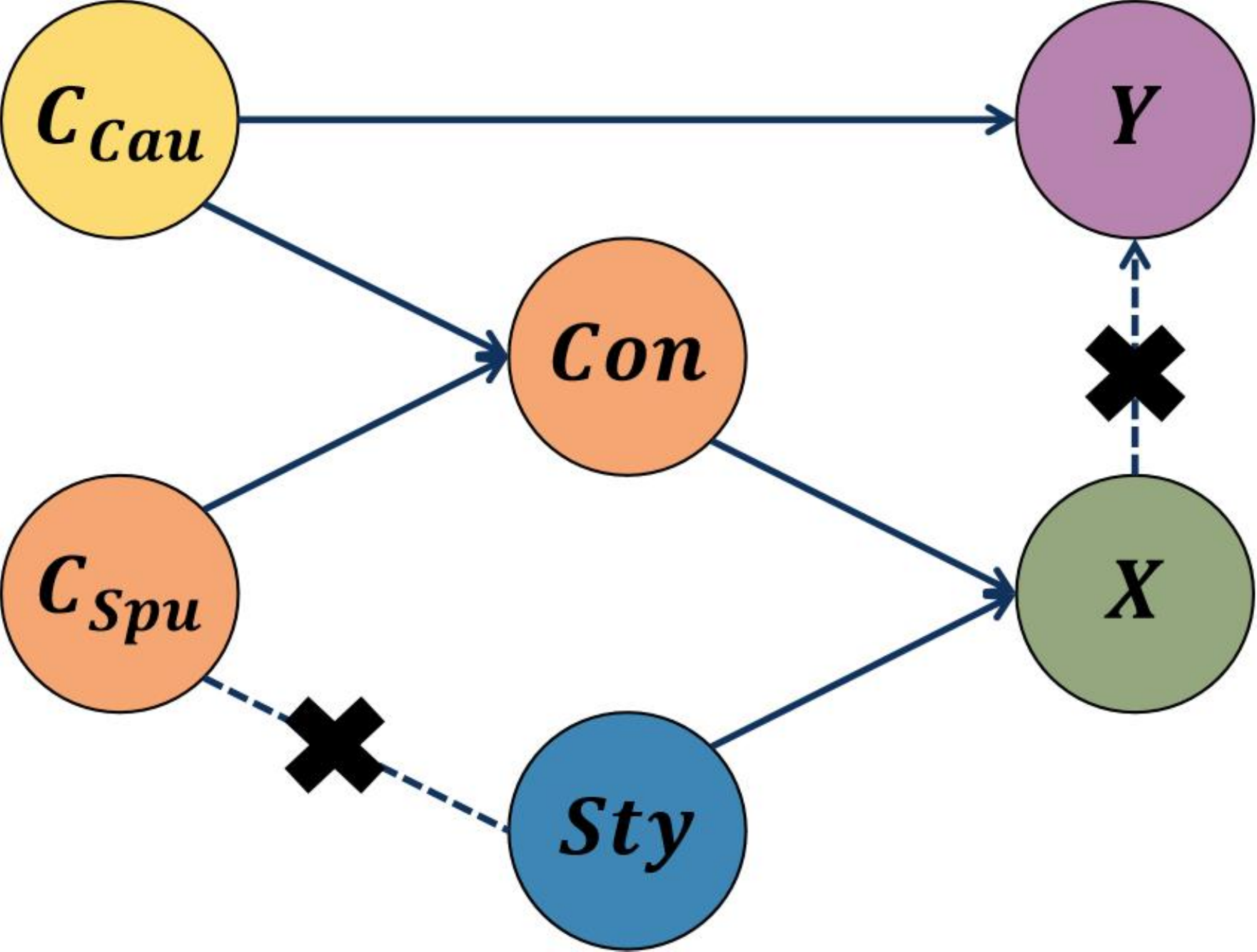} 
    \caption{The proposed SCM for image generation, which serves as the causal-inspired assumption of IRSS. The causal features of the target object in the image are represented by $C_{cau}$. In contrast, the unrelated features are represented by $C_{spu}$. The combination of $C_{cau}$ and $C_{spu}$ forms the image's overall feature, denoted as content $Con$. $Con$ is then combined with the style feature $Sty$ to generate the final image $X$. Additionally, $C_{cau}$ determines the classification label $Y$.}
    \label{fig:IRSSSCM}
\end{figure}

Inspired by the above observations on the essential factors of OOD, we further conduct in-depth image analysis (including a series of experiments using Segment Anything \cite{kirillov2023segment}). This in-depth image analysis enables us to formulate a new Structural Causal Model (SCM) for image generation that explicitly captures style and spurious features, as shown in Figure \ref{fig:IRSSSCM}. There is no direct causal relationship between the image and the label, nor between the spurious feature and the style (although a correlation may exist in some datasets).  

Based on the above proposed new SCM model for image generation, we propose a new framework, \textbf{IRSS} (\underline{I}nvariant \underline{R}epresentation Learning via Decoupling \underline{S}tyle and \underline{S}purious Features). IRSS aligns the style distribution and eliminates the influence of spurious features using two independent components. Unlike previous methods that require domain labels, our approach requires two pieces of information-style domain division and environmental division. Our method only requires image features and labels and does not need any additional information. And our method achieves excellent performance on the dataset PACS, OfficeHome, and NICO, demonstrating that our approach can learn the invariance well from images.

\section{Related Works}

To deal with domain shift \cite{torralba2011unbiased}, domain adaptation and domain generalization have been studied extensively. Both domain adaptation and domain generalization address the problem of transferring knowledge from a source domain to a target domain when they generate datasets using different probability distributions.

Domain generalization problems are generally more challenging than domain adaptation problems. Our work aims to tackle the difficult task of domain generalization, and it is robust to erroneous or even missing domain labels. Methodologically, our work is closely related to two research lines: (1) the domain adversarial neural network line and (2) the invariant risk minimization line. 


\paragraph{The domain adversarial neural network line}  Ganin et al. proposed the domain adversarial neural network (DANN) \cite{ganin2016domain}, which is a representative domain adaptation framework. It learns domain-invariant features by inducing a domain regularizer into the training process. DANN inspired a large number of works. Notable works in domain adaptation include \cite{long2015learning, ganin2016domain, tzeng2017adversarial}. As the field of domain generalization continues to expand, works such as \cite{fan2021adversarially, matsuura2020domain} applied the ideas of DANN to this area. However, DANN does not assume the existence of spurious features separately but integrates them with the style distribution. This integration of features is effective on datasets with specific correlations between spurious features and style features, e.g., isolating the target object in the blank space in sketches. It may lead to poor performance on datasets with weak correlation between spurious features and style features. Our method separates the style features and spurious features. Motivated by \cite{ganin2016domain, matsuura2020domain}, we incorporate a component of adversarial networks to align the style distributions only.  
Through this, we address the limitation of DANN.  

\paragraph{The invariant risk minimization line} Arjovsky et al. proposed the Invariant risk minimization (IRM) framework \cite{arjovsky2019invariant}. Based on the work \cite{peters2016causal}, IRM applies SCM to deep learning, which regularizes neural networks to extract invariant features and discard spurious features. Specifically, IRM considers the setting that the training data is collected from multiple environments where the correlation between spurious features and labels varies across environments. It assumes that the correlation of invariant features remains stable. IRM inspired a large number of follow-up works. Many works extended the penalty functions. InvRat \cite{chang2020invariant} developed an entropy-based adversarial penalty function. REx \cite{krueger2021out} developed a variance-based adversarial penalty function. Other notable works that extended the penalty functions of IRM include PAIR \cite{chen2022pareto}, BIRM \cite{lin2022bayesian} and SIRM \cite{zhou2022sparse}. Although IRM was shown to have superior performance in various applications, recent works such as \cite{gulrajani2020search, lin2021empirical, rosenfeld2020risks} found that IRM has poor performance on deep models. It performs poorly when there is a significant distribution shift in the dataset. By introducing adversarial networks, our approach achieves distribution-aligned feature maps, which enables IRM regularization to extend its scope, even for datasets with significant distribution shifts. 

\begin{table}[htb]
\resizebox{\columnwidth}{!}{
\begin{tabular}{lcc}
\hline
\textbf{Methods} &\emph{causal $\|$ Style} & \emph{causal $\|$ Spurious} \\
\hline 
Domain Generalization \\
DANN~\cite{ganin2016domain} & \ding{51} & \ding{55} \\
Cumix~\cite{mancini2018best} & \ding{51} & \ding{55} \\
MMLD~\cite{matsuura2020domain} & \ding{51} & \ding{55} \\
DecAug~\cite{bai2021decaug} & \ding{51} & \ding{51} \\
JiGen~\cite{carlucci2019domain} & \ding{55} & \ding{51} \\
StableNet~\cite{stablenet} & \ding{55} & \ding{51} \\
\hline
\hline
Invariant Risk Minimization \\
IRM~\cite{rosenfeld2020risks} & \ding{55} & \ding{51} \\
InvRat\cite{chang2020invariant} & \ding{55} & \ding{51} \\
REx~\cite{krueger2021out} & \ding{55} & \ding{51} \\
BIRM~\cite{lin2022bayesian} & \ding{55} & \ding{51} \\
SIRM~\cite{zhou2022sparse} & \ding{55} & \ding{51} \\
PAIR~\cite{chen2022pareto} & \ding{55} & \ding{51} \\
IIB~\cite{li2022invariant} & \ding{55} & \ding{51} \\
\hline
\hline 
IRSS(\textbf{Ours}) & \ding{51} & \ding{51}  \\
\hline 
\end{tabular}
}
\caption{
Summary of existing works on out-of-distribution generalization.
\emph{causal $\|$ Style} and \emph{causal $\|$ spurious} mean that methods achieve disentangling the causal features with style features and spurious features, respectively.
}
\label{tab:related_work}
\end{table}



\section{Model and Design Overview of IRSS}

\subsection{Model}


We consider a training dataset denoted by $\mathcal{D}_{tr} = \{(\boldsymbol{x}_i, y_i)\}^N_{i=1}$, where $\boldsymbol{x}_i \in \mathbb{R}^D$ denotes the raw feature vector with $D\in \mathbb{N}_+$, $y_i \in \mathcal{C}\triangleq \{1,\ldots,C\}$ denotes the label with $ C\in \mathbb{N}_+$, and $N \in \mathbb{N}_+$ denotes the number of data points.  We consider the setting that the data points in the training dataset may be generated from different domains, and the domain labels are unknown.  
Let $\mathcal{E}_{tr}$ denote the set of domains or environments that generate the training datasets.  
Let $F_f : \mathbb{R}^D \rightarrow \mathbb{R}^d$ denote a feature map or feature extractor parameterized by $\boldsymbol{\theta}_f$, 
where $d \in \mathbb{N}_+$. Let $F_y: \mathbb{R}^d \rightarrow \Delta(\mathcal{C})$ denote a label predictor parameterized by $\boldsymbol{\theta}_y$, where $\Delta(\mathcal{C})$ denotes a probability simplex over $\mathcal{C}$:  
\[
\Delta(\mathcal{C})
\triangleq 
\{
\boldsymbol{p} \in \mathbb{R}^C | 
p_1 + \cdots + p_C =1, 
p_c \geq 0, \forall c \in \mathcal{C}
\}.
\]  
Namely, $F_y$ outputs a probability vector that characterizes the probability mass function of the predicted label. Note that $F_y$ captures the deterministic prediction function as a special case 
of unit probability vectors.  
For simplicity of presentation, let $F_y(\cdot)_c$ denote the probability that the predicted label is $c \in \mathcal{C}$.  
The loss function associated with the label classifier $F_y$ is denoted as $\ell_{\operatorname{ERM}}$ and 
expressed as follows:
\begin{equation}
\begin{aligned}
\ell_{\operatorname{ERM}}(\!\boldsymbol{\theta}_f, 
\boldsymbol{\theta}_y; \mathcal{D}_{tr} \!) 
{=}
{-}  \frac{1}{N} \!\! \sum_{i=1}^{N} \! 
\log \!F_y\! \left(F_f \!\left(\boldsymbol{x}_i; \boldsymbol{\theta}_f \!\right); 
\boldsymbol{\theta}_y \right)_{y_i}.
\end{aligned}
\label{eq:ERMloss}
\end{equation} 
Our objective is to learn $F_f$ and $F_y$ such that the predictor can generalize well across all (possibly unseen) domains.  

\subsection{Design Overview of IRSS}

\begin{figure}[htb]
    \centering
    \includegraphics[width=\columnwidth]{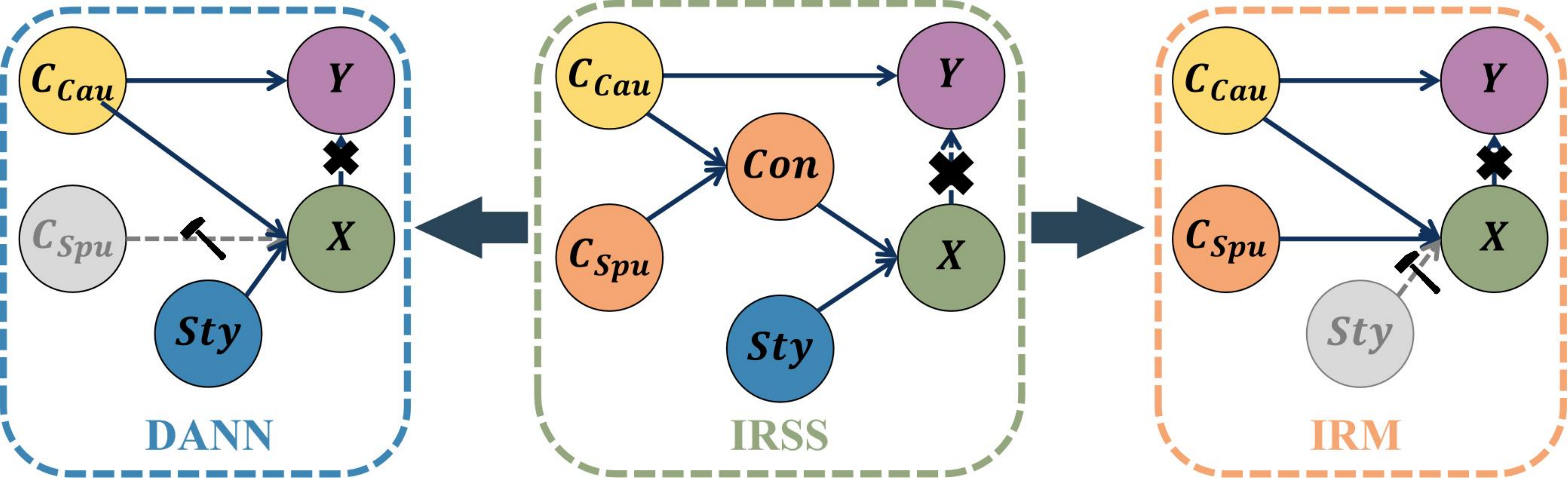}
    \caption{IRSS degenerates into DANN and IRM}
    \label{fig:formprevious}
\end{figure}

\paragraph{Structural causal model of IRSS.} Figure \ref{fig:formprevious} depicts the structural causal model of IRSS. 
The SCM treats style distribution shift and spurious feature distribution shift as two separate confounders.  
Technically, this SCM has two potential merits: 
(1) addresses the lack of the ability to separate spurious features in DANN; 
(2) resolves the failure of IRM in the case of significant distribution shifts. 

\begin{figure*}[htb]
    \centering
    \includegraphics[width=.9\textwidth]{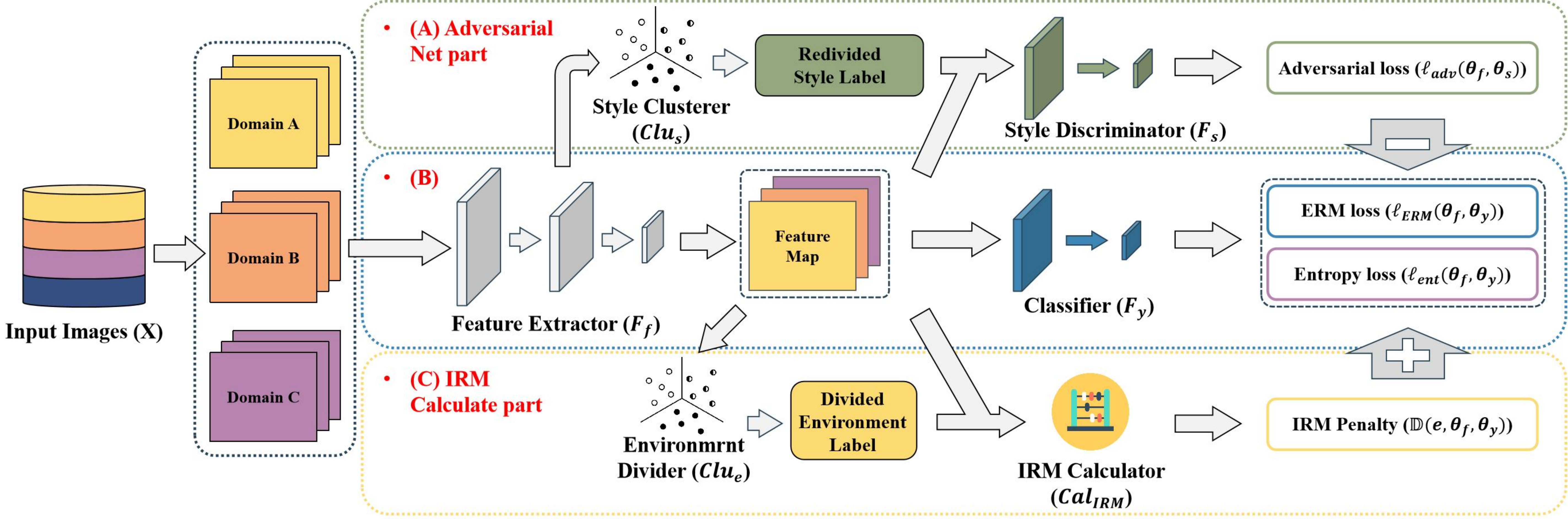} 
    \caption{The proposed method framework \textbf{IRSS} consists of three main parts: (A) Adversarial Net Part, (B) Main Network, and (C) IRM Calculate Part.}
    \label{fig:IRSSframework}
\end{figure*}

We elaborate more on the first potential merit of the proposed SCM. Figure \ref{fig:formprevious} depicts the SCM of the DANN method. This SCM posits that the image $X$ is determined solely by the causal features (domain invariant) and style features. It does not model the spurious features explicitly; instead, it captures the spurious features implicitly in the style features. This model of spurious features is effective on datasets with specific correlations between spurious features and style features, e.g., isolating the target object in the blank space in sketches. It may lead to poor performance on datasets where the correlation between spurious features and style features is weak. For example, as illustrated in Figure \ref{oodexample}, even in a cartoon style, elephants are in various environments, and DANN has no mechanism to remove the influence of spurious features.  

We elaborate more on the second potential merit of the proposed SCM. Figure \ref{fig:formprevious} also depicts the SCM of IRM-like methods. In contrast to our proposed approach, this model assumes the independent existence of spurious features. According to this model, the image $X$ is determined by both the causal and spurious features. To better understand it, we consider IRMv1 \shortcite{arjovsky2019invariant}, which assumes only linear $F_f$ and $F_y$. Referring to the SCM of IRM shown in Figure \ref{fig:formprevious}, let's denote the causal features and spurious features as $\boldsymbol{z}_c$ and $\boldsymbol{z}_e$ respectively. Then, we can express $\boldsymbol{x}$ as a function of $\boldsymbol{z}_c$ and $\boldsymbol{z}_e$ as $\boldsymbol{x}=g_1(\boldsymbol{z}_c,\boldsymbol{z}_e)$, and $\boldsymbol{y}$ as a function of $\boldsymbol{z}_c$ as 
$\boldsymbol{y}=g_2(\boldsymbol{z}_c)$, where both $g_1$ and $g_2$ are set to be linear functions. Then an ideal $F_f$ and $F_y$ are as follows:  
\begin{equation}
\begin{aligned}
F_f(\boldsymbol{x})=\left[\begin{array}{ll}
I & 0 \\
0 & 0
\end{array}\right] &\circ g^{-1}_1(\boldsymbol{x})=[\boldsymbol{z}_c]\\
F_y(F_f(\boldsymbol{x}))&=g_2(\boldsymbol{z}_c).
\end{aligned}
\label{eq:IRM_ideal}
\end{equation}

However, as stated in \cite{rosenfeld2020risks}, IRM performs poorly when there is a significant distribution shift in the dataset. Their theoretical results demonstrate that the 0-1 risk of classical IRM would suffer a lower bound relying on a Gaussian CDF and a ratio of the variances between test and training datasets.  






\paragraph{Framework of IRSS.}  
Figure \ref{fig:IRSSframework} shows the overall framework of IRSS, which consists of three components:

\begin{enumerate}
    \item \textbf{Aligning style distribution}: This component 
    involves clustering the features extracted by a shallow feature extractor to obtain style labels. It utilizes the ``(A) Adversarial Net Part'' to get alignment features independent of the style distribution.

    \item \textbf{Eliminating the influence of spurious features}: In this component, the aligned features are re-clustered to partition the features into different environments. The ``(C) IRM Calculate Part'' calculates the penalty term associated with these features.  

    \item \textbf{Loss Calculation and training}: The final loss is obtained by weighting and summing the empirical risk minimization loss and the Entropy loss brought by the ``(B) main network''. The Adversarial and IRM penalty losses are incorporated into the overall loss calculation. The model is trained based on this composite loss function.
\end{enumerate}

In the following section, we design algorithms to implement the above framework.

\section{The Algorithm for IRSS}

\subsection{Aligning Style Distribution}

\paragraph{Extracting style-Specific distinguishing features.} We aim to extract invariant features across different styles using adversarial neural networks. However, utilizing style domain labels directly in DANN often results in several challenges: 
1): missing or inaccurate labels during data collection, 
2): capturing partial style distribution differences in the labels, and 
3): varying similarities between different styles (e.g., sketches and art are more similar to each other than photos).

To achieve enhanced alignment performance and provide more consistent distributions, we adopt the style features in style transfer \cite{li2017demystifying} as the domain discriminative features \cite{matsuura2020domain}. We use convolutional features as the domain discriminative features based on the assumption that image domains are reflected in their style distributions. We define the stack of multi-scale style features obtained from different convolutional layers as the style discriminative features. The style discriminative features, $\operatorname{sdf}(\boldsymbol{x})$, are computed using the multi-layer outputs, $\phi_1(\boldsymbol{x}),\ldots,\phi_M (\boldsymbol{x})$. Here, $M\in \mathbb{N}_+$ denotes the number of output layers employed, while $\mu(\cdot)$ and $\sigma(\cdot)$ represent the mean and variance functions.
\begin{equation}
\begin{aligned}
\operatorname{sdf}(\boldsymbol{x})
=
\left(
\mu\left(\phi_m(\boldsymbol{x})\right); \sigma\left(\phi_m(\boldsymbol{x})\right) : 
m = 1,\ldots,M
\right).
\end{aligned}
\label{eq:sdf}
\end{equation}

After obtaining the style discriminative features, 
$\operatorname{sdf}(\boldsymbol{x})$, we employ standard clustering methods to re-divide them into style domain labels. Let $\hat{s}_i \in \mathcal{S}$ denote the style label assigned to data $\boldsymbol{x}_i$, 
where $\mathcal{S} \triangleq \{1,\ldots,S\}$ denotes a set of all possible styles.   In our experiments, we utilize the K-means algorithm for this purpose.

\paragraph{Minimizing adversarial loss.}

Inspired by DANN, we introduce a style discriminator $F_s$ to our model. Formally, $F_s: \mathbb{R}^d \rightarrow \Delta(\mathcal{S})$ is parameterized by $\boldsymbol{\theta}_s$, where $\Delta(\mathcal{S})$ denotes a probability simplex over $\mathcal{S}$: 
\[
\Delta(\mathcal{S})
\triangleq 
\{
\boldsymbol{p} \in \mathbb{R}^S | 
p_1 + \cdots + p_S =1, 
p_s \geq 0, \forall s \in \mathcal{S}
\}.
\] 
Namely, $F_s$ outputs a probability vector that characterizes the probability mass function of the predicted style. For the simplicity of presentation, let $F_s(\cdot)_s$  denote the probability that the predicted label is $s \in \mathcal{S}$. The $F_s$ is trained to distinguish styles based on the output of the feature extractor. In contrast, the feature extractor is trained to extract style-invariant features, making it difficult for the style discriminator to differentiate between styles. This enables the extraction of style-invariant features from multiple source styles, which summarizes the model's capacity to generalize to unseen target styles. The adversarial loss $\ell_{\operatorname{adv}}$ is defined as follows: 
\begin{equation}
\begin{aligned}
\ell_{\operatorname{adv}}
(\boldsymbol{\theta}_f,\boldsymbol{\theta}_s)=-\frac{1}{N} \sum_{i=1}^{N} 
\log F_s\left(F_f\left(\boldsymbol{x}_i; 
\boldsymbol{\theta}_f \right); 
\boldsymbol{\theta}_s \right)_{\hat{s}_i}. 
\end{aligned}
\label{eq:advloss}
\end{equation}  

Furthermore, we introduce entropy loss $\ell_{\operatorname{ent}}$ (Grandvalet and Bengio 2005) to our model. It is utilized in some domain adaptation methods (Long et al. 2016; Zhang et al. 2019) to train a more discriminative model for target samples by encouraging low-density separation among object categories. 
\begin{equation}
\begin{aligned}
\ell_{\operatorname{ent}}
(\boldsymbol{\theta}_f,\boldsymbol{\theta}_y) 
=
-\frac{1}{N} \sum_{i=1}^{N} H\left(F_y\left(F_f\left(x_i;
\boldsymbol{\theta}_f
\right);
\boldsymbol{\theta}_y
\right)\right).
\end{aligned}
\label{eq:entloss}
\end{equation}

\subsection{Eliminating the Influence of Spurious Features}

Based on the alignment of style distribution completed in the previous stage, we partition samples with similar spurious features into the same environment by leveraging the obtained feature map. This excludes the influence of incorrect features by transforming the problem into a multi-environment learning problem. We again use clustering algorithms to partition the feature map and obtain multiple environments. Let $\hat{\mathcal{E}}_{tr}$ denote a set of all the obtained environments. Let $e_i \in \hat{\mathcal{E}}_{tr}$ denote the environment label of data $(\boldsymbol{x}_i, y_i)$. For each environment label $e \in \hat{\mathcal{E}}_{tr}$, we defined the associated sub-set of training dataset as: 
\[
\mathcal{D}_e 
\triangleq 
\{
(x_i,y_i) 
:  
e_i = e, i =1,\ldots,N
\}.  
\]

Subsequently, we adopt the idea of IRM to learn across multiple environments: we aim to learn a model that exhibits good generalization across all environments, which we call invariance. 
The optimization objective of our method can be formulated as follows, where $\bar{\boldsymbol{\theta}}_f$ and $\bar{\boldsymbol{\theta}}_y$ denote the Pareto fronts of $\boldsymbol{\theta}_f$ and $\boldsymbol{\theta}_y$, respectively. 
\begin{equation}
\begin{aligned}
\min_{\boldsymbol{\theta}_f, 
\boldsymbol{\theta}_y } & \ \  \left[\sum_{e\in\mathcal{E}_{tr}}  
\ell_{\operatorname{ERM}} 
(\boldsymbol{\theta}_f, \boldsymbol{\theta}_y;
\mathcal{D}_e)
\right] +\lambda_{\operatorname{ent}}\cdot \ell_{\operatorname{ent}} 
(\boldsymbol{\theta}_f,\boldsymbol{\theta}_y)\\ 
-\lambda_{\operatorname{adv}} & \cdot \ell_{\operatorname{adv}} 
(\boldsymbol{\theta}_f,\boldsymbol{\theta}_s)
\quad\text{s.t.}\ \boldsymbol{\theta}_s \in 
\arg\min_{\boldsymbol{\theta}_s}\ell_{\operatorname{adv}}(\boldsymbol{\theta}_f,\boldsymbol{\theta}_s) \\
 & \ \  \boldsymbol{\theta}_f,
\boldsymbol{\theta}_y\in
\underset{\bar{\boldsymbol{\theta}}_f,\bar{\boldsymbol{\theta}}_y}{\arg\min}\ell_{\operatorname{ERM}}
(\bar{\boldsymbol{\theta}}_f,\bar{\boldsymbol{\theta}}_y;
\mathcal{D}_e),\ \forall e\in \hat{\mathcal{E}}_{tr}.
\end{aligned}
\label{eq:overall_problem}
\end{equation}

To address this bi-level optimization problem, we leverage the idea of Lagrange multipliers. It introduces a differentiable $\mathbb{D}(e,\boldsymbol{\theta}_f, \boldsymbol{\theta}_y)$ concerning $\boldsymbol{\theta}_f$ and $\boldsymbol{\theta}_y$, which is subject to the following minimization constraint as much as possible: 
\begin{equation}
\begin{aligned}
\min_{\boldsymbol{\theta}_f, 
\boldsymbol{\theta}_y} & \ \  \left[\sum_{e\in \hat{\mathcal{E}}_{tr}}  
\ell_{\operatorname{ERM}} 
(\boldsymbol{\theta}_f, \boldsymbol{\theta}_y;
\mathcal{D}_e) 
+ \lambda_{\operatorname{IRM}}\cdot\mathbb{D}(e,\boldsymbol{\theta}_f, \boldsymbol{\theta}_y)
\right]\\ & +\lambda_{\operatorname{ent}}\cdot \ell_{\operatorname{ent}} 
(\boldsymbol{\theta}_f,\boldsymbol{\theta}_y)
-\lambda_{\operatorname{adv}} \cdot \ell_{\operatorname{adv}} 
(\boldsymbol{\theta}_f,\boldsymbol{\theta}_s)
\\
&\qquad\text{s.t.} \ \ 
\boldsymbol{\theta}_s \in 
\arg\min_{\boldsymbol{\theta}_s}  \ \ \ell_{\operatorname{adv}}(\boldsymbol{\theta}_f,\boldsymbol{\theta}_s).
\end{aligned}
\label{eq:finalform}
\end{equation}

In our experiments, we used two forms of IRM penalty $\mathbb{D}(e,\theta_f, \theta_y)$, namely IRMv1 \cite{arjovsky2019invariant} and BIRM \cite{lin2022bayesian}, which are calculated as follows for a specific environment $e$, 
\begin{equation}
\begin{aligned}
&\mathbb{D}^e_{\operatorname{IRMv1}}= \left\|\nabla_{w\mid w=1.0}  \ell_{\operatorname{ERM}} 
( \boldsymbol{w} \cdot \boldsymbol{\theta}_f, 1;
\mathcal{D}_e)\right\|^2\\
&\mathbb{D}^e_{\operatorname{BIRM}} = \sum_{(x_i,y_i)\in \mathcal{D}_e}\left[\ln p\left(y_i \mid \boldsymbol{\theta}_f,\boldsymbol{\theta}_y^e\right) -\ln p\left(y_i \mid \boldsymbol{\theta}_f,\boldsymbol{\theta}_y\right)\right].
\end{aligned}
\label{eq:IRMpenalties}
\end{equation}
where $w$ is a dummy variable set to 1 and $\boldsymbol{\theta}^e_y$ is the best classifier parameter for $F_y$ on environment $e$. 

\subsection{Overall algorithm.}  
Putting them together, Algorithm \ref{alg:IRSS} outlines our detailed algorithmic procedures of IRSS. For training efficiency consideration, we update the estimation of style labels infrequently, i.e., in big steps only. Furthermore, we sample a mini-batch of the training dataset in each step to train model parameters. 
 
\begin{algorithm}[h]
\caption{Invariant Representation Learning via Decoupling Style and Spurious Features}
\label{alg:IRSS}
\textbf{Input}: $\mathcal{D}=\left\{x_i, y_i\right\}_{i=1}^{N}, \boldsymbol{\theta}_f^{(0)},\boldsymbol{\theta}_y^{(0)}, \boldsymbol{\theta}_s^{(0)}$, $\operatorname{iter}=0$, training parameters\\
\textbf{Output}: $\boldsymbol{\theta}_{f}^{(out)},\boldsymbol{\theta}_{y}^{(out)}$
\begin{algorithmic}[1] 
\STATE Initialize $\{s_i\}_{i=1}^N,\{e_i\}_{i=1}^N$ with zero
\WHILE{not end of bigstep}
\STATE Calculate $\left\{\operatorname{sdf}\left(x_i\right)\right\}_{i=1}^{N}$ using Eq.(\ref{eq:sdf})
\STATE Update $\{\hat{s}_i\}_{i=1}^N$ by clustering $\left\{\operatorname{sdf}\left(x_i\right)\right\}_{i=1}^{N}$
\WHILE{not end of step}
\STATE Sample a minibatch of $\{x_i^m, y_i^m, \hat{s}_i^m\}$
\STATE Update $\{e_i^m\}$ by clustering $F_f(x_i^m;\boldsymbol{\theta}_f^{(\operatorname{iter})})$
\STATE Update $\boldsymbol{\theta}_f^{(\operatorname{iter+1})}, \boldsymbol{\theta}_y^{(\operatorname{iter+1})}, \boldsymbol{\theta}_s^{(\operatorname{iter+1})}$ using Eq.(\ref{eq:finalform})
\STATE $\operatorname{iter} = \operatorname{iter}+1$
\ENDWHILE
\ENDWHILE
\end{algorithmic}
\end{algorithm}




\section{Experiments}

\begin{table*}[htb]
\centering
\begin{tabular}{lcccc|c|cccc|c}
\hline
\textbf{Dataset} & \multicolumn{5}{c|}{\textbf{PACS}} & \multicolumn{5}{c}{\textbf{OfficeHome}}\\
\hline
\textbf{Domain}   &\textbf{Art} & \textbf{Cartoon}     & \textbf{Photo}       & \textbf{Sketch}      & \textbf{Avg}   & \textbf{Art} & \textbf{Clipart} & \textbf{Product} & \textbf{RealWorld} & \textbf{Avg}\\
\hline
ERM                                             & 77.85     & 74.86     & 95.73     & 67.74     & 79.05     & 49.83     & 48.26     & 67.21     & 67.59     & 58.22 \\
IRM           & 70.31     & 73.12     & 84.73     & 75.51     & 75.92     & 41.88     & 36.01     & 55.87     & 60.54     & 48.57 \\
D-SAM-$\Lambda$        & 79.48     & 77.13     & 94.30     & 75.30     & 81.55     & 54.53     & 49.04     & 71.57     & 71.90     & 61.76 \\
REx                 & 67.04     & 67.97     & 89.74     & 59.81     & 71.14     &  52.87         & 44.56          & 67.62          &  70.36         & 58.85      \\
DecAug               & 81.33     & 78.39     & 93.71     & 77.17     & 82.65     & 51.61     & 48.00     & 69.19     & 69.63     & 59.61 \\
IIB                & 77.02     & 73.02     & 88.76     & 75.82     & 78.76     & 35.46         & 38.21         & 54.21         & 54.16         & 45.51       \\
JiGen           & 79.42     & 75.25     & \underline{96.03}     & 71.35     & 80.51     & 53.04     & 47.51     & 71.47     & 72.79     & 61.20 \\
MMLD            & 81.28     & 77.16     & \textbf{96.09}     & 72.29     & 81.83     & 52.32     & 45.45     & 68.56     & 70.71     & 59.26 \\
Cumix           & 82.30     & 76.50     & 95.10     & 72.60     & 81.60     & -         & -         & -         & -         & -     \\
DDAIG                & \textbf{84.20}     & 78.10     & 95.30     & 74.70     & 83.10     &  \underline{58.66}   & 43.16   & 70.96   & 74.24    & 61.76  \\

\hline
\textbf{IRSS-IRMv1} (Ours)                     & 82.29     & \textbf{81.20}     & 94.07     & \underline{78.70}     & \underline{84.07}       &   58.44      &       \underline{49.43}    &  \underline{72.78}         &  \textbf{75.57}         &  \underline{64.06}     \\
\textbf{IRSS-BIRM}  (Ours)                     & \underline{82.74}     & \underline{80.51}     & 94.73     & \textbf{79.54}     & \textbf{84.38}     &      \textbf{59.56}     &  \textbf{49.73}     &  \textbf{73.80}      &  \underline{75.32}         &  \textbf{64.60}     \\
\hline
\end{tabular}

\caption{Classification accuracy of our approach trained considering leave-one-domain-out validation compared with the state-of-the-art methods on the PACS and OfficeHome benchmark with the ResNet-18 backbone. The open-source code of Cumix does not support running on OfficeHome.}
\label{tab:CompareMain}
\end{table*}

\subsection{Experimental Setting}


\paragraph{Dataset.} 
We conducted experiments on three challenging and representative datasets, i.e., 
PACS \cite{li2017deeper}, OfficeHome \cite{venkateswara2017deep}, and NICO \cite{he2021towards}, all of which exhibit varying style distributions and non-target content distributions. 

The PACS dataset \cite{li2017deeper} consists of four domains, \textbf{Art} Painting, \textbf{Cartoon}, \textbf{Photo} and \textbf{Sketch}, each containing seven common categories.
The OfficeHome dataset \cite{venkateswara2017deep} consists of four domains; \textbf{Art}, \textbf{Clipart}, \textbf{Product} and \textbf{RealWorld}, each containing 65 different categories.
The NICO dataset \cite{he2021towards} contains 19 classes with 9 or 10 different contexts, i.e., different object poses, positions, backgrounds, movement patterns, etc. Compared to traditional datasets, NICO provides a better categorization of non-target content.  
The primary task of IRSS is to perform classification on unseen datasets. For PACS and OfficeHome, the original domain split was used as the basis for domain labels, and we adopted the same leave-one-domain-out cross-validation protocol as in \cite{li2017deeper}. We merge different attributes for NICO to form three training domains, one validation domain, and one test domain. 
Table \ref{tab:datasets} shows the number of images in each dataset.  

\begin{table}[htb]
\resizebox{\columnwidth}{!}{
    \centering
    \begin{tabular}{lc|lc|lc|c}
    \hline
        \multicolumn{2}{c|}{\multirow{2}{*}{\textbf{PACS}}} & \multicolumn{2}{c|}{\multirow{2}{*}{\textbf{OfficeHome}}} &  \multicolumn{3}{c}{\textbf{NICO}} \\ \cline{5-7}
        ~ & ~ & ~ & ~ & \textbf{Subdataset} & \textbf{Animal} & \textbf{Vehicle} \\ \hline
        Art & 2048 & Art & 2427 & Domain1 & 3046 & 2379 \\ 
        Cartoon & 2344 & Clipart & 4365 & Domain2 & 4154 & 3737 \\ 
        Photo & 1670 & Product & 4439 & Domain3 & 3913 & 3100 \\ 
        Sketch & 3929 & RealWorld & 4357 & Validation & 843 & 1079 \\ 
        - & - & - & - & Test & 1093 & 1355 \\ \hline
        \textbf{Overall} & 9991 & \textbf{Overall} & 15588 & \textbf{Overall} & 13049 & 11650 \\ \hline
    \end{tabular}
}
\caption{Each domain in PACS, OfficeHome, and NICO datasets and its corresponding number of pictures.}
\label{tab:datasets}
\end{table}

\paragraph{Metrics.} 
The performance metric is the top-1 category classification accuracy. We use a ResNet-18 backbone for all datasets. The data of each training domain was split randomly into 80\% for training and 20\% for validation.

\noindent
\paragraph{Baselines.} 
We compare our method with the following recent domain generalization methods. 
{\bf ERM:} Pre-trained ResNet-18 is fine-tuned on the aggregated data from all source domains, using only the classification loss. 
\textbf{IRM} \cite{arjovsky2019invariant}: Add IRMv1 penalty term to the classification loss of ERM. 
\textbf{D-SAM-$\Lambda$} \cite{d2019domain}: Introduce domain-specific aggregation modules. 
\textbf{REx} \cite{krueger2021out}: Use variance as a penalty. 
\textbf{JiGen} \cite{carlucci2019domain}: A puzzle-based generalization method on unsupervised tasks solving jigsaw puzzles. 
\textbf{MMLD} \cite{matsuura2020domain}: Attempt to use a mixture of multiple latent domains as novel and more realistic scenarios for domain generalization. 
\textbf{CuMix} \cite{mancini2020towards}: Aim to address the test-time domain and semantic shift by simulating them by utilizing images and features from unseen domains and categories. Its open-source code does not support running on OfficeHome. 
\textbf{DDAIG} \cite{zhou2020deep}: A method based on deep domain adversarial image generation. 
\textbf{DecAug} \cite{bai2021decaug}: Apply semantic augmentation after decomposing features. 
\textbf{IIB} \cite{li2022invariant}: Aim to minimize the invariance risk of non-linear classifiers while alleviating the impact of spurious invariant features and geometric bias.

During the comparison process, we standardized all methods by uniformly implementing a random seed-based proportional random split.

\subsection{Comparison with Baselines}

We evaluated our method on PACS, OfficeHome, and NICO, whose results are presented in Tables \ref{tab:CompareMain} and \ref{tab:NICO}.  
One can observe that our two approaches outperform all baselines and achieve the best performance across all datasets.

\begin{table}[htb]
\centering
\resizebox{0.75\columnwidth}{!}{
\begin{tabular}{lc|c}
\hline
\textbf{Dataset} & \multicolumn{2}{c}{NICO} \\
\hline
\textbf{Subdataset}   & \textbf{Animal}      & \textbf{Vehicle}     \\
\hline
ERM                                      & 75.87                & 74.52                \\
IRM    & 59.17                & 62.00                \\
REx         & 74.31                & 66.20                \\
Cumix    & 76.78                & 74.74                \\
DecAug     & 85.23                & 80.12                \\
IIB        & 88.17  &     81.57                     \\
JiGen    & 84.95                & 79.45                \\
MMLD     & 74.60 & 71.88                      \\
\hline
\textbf{IRSS-IRMv1} (Ours)              &    \textbf{91.37}  & \underline{83.95} \\
\textbf{IRSS-BIRM}  (Ours)              &    \underline{91.21}  & \textbf{84.06} \\
\hline
\end{tabular}}
\caption{Results of our method compared with different models on the NICO dataset.}
\label{tab:NICO}
\end{table}

Although our method generally outperforms the baselines on most of the three datasets, it could have performed better in the sub-experiment of PACS, precisely when Photo was used as the testing domain. After carefully observing and analyzing PACS, 
we explain that Photos in PACS contains more spurious features than other style domains. Despite our efforts to mitigate their influence during training by partitioning the spurious features, there are still newly emerging spurious features during testing that our method results in suboptimal performance.

\subsection{Sensitivity of Environment Nums and Style Nums}

\begin{figure}[ht]
    \centering
    \subfloat[Sensitivity Analysis on PACS Dataset]{\includegraphics[width=\columnwidth]{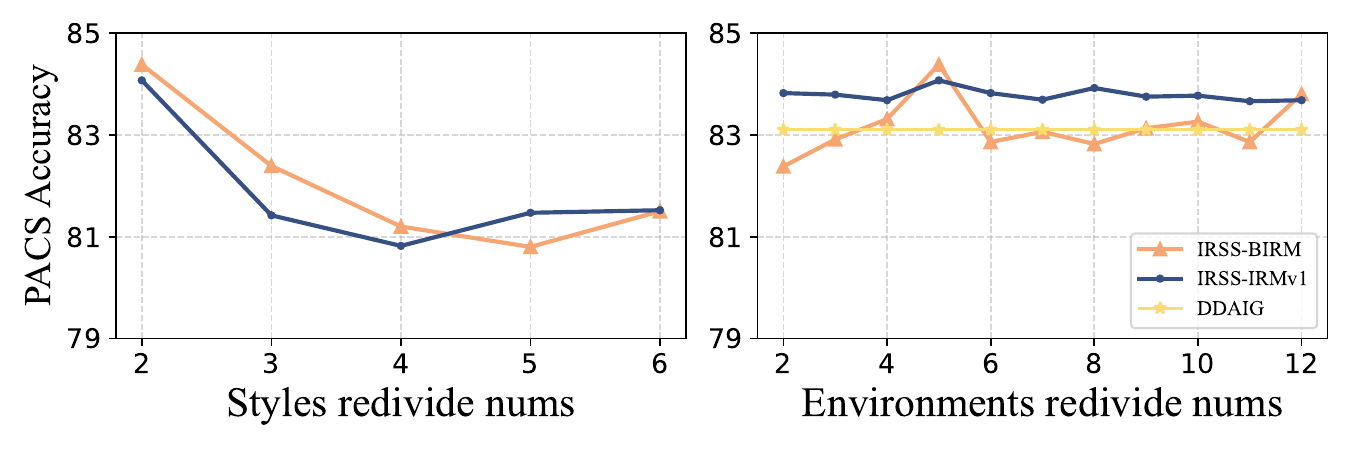}
    \label{fig:sensitivity:PACS}
    }\\
    \subfloat[Sensitivity Analysis OfficeHome Dataset]{\includegraphics[width=\columnwidth]{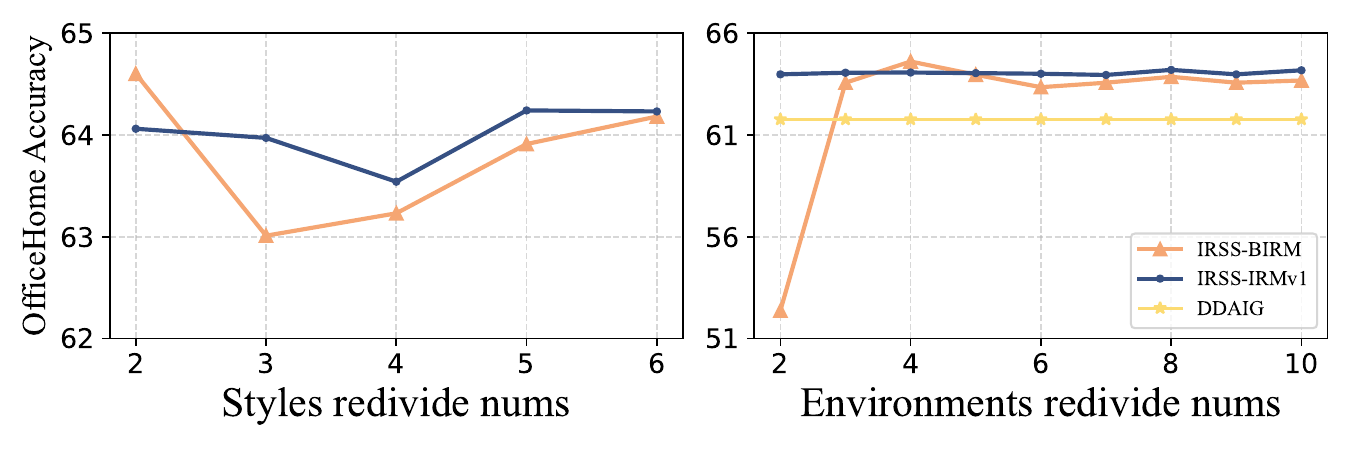}
    \label{fig:sensitivity:OfficeHome}
    }
    \caption{Sensitivity Analysis on Environment Nums and Style Nums in the PACS and OfficeHome Datasets}
    \label{fig:sensitivity}
\end{figure}

We perform sensitivity analysis on the 
hyper parameter of style and environment division,  i.e., number of styles and number environments, as shown in Figure \ref{fig:sensitivity}. 
Figure \ref{fig:sensitivity:PACS}
presents the sensitivity analysis on the PACS dataset. We fix the number of environment divisions to 5 to investigate the influence of style division on performance and fix the number of style divisions at 2 to examine the impact of environment division on performance. 
Figure \ref{fig:sensitivity:OfficeHome} 
displays the sensitivity analysis on the OfficeHome 
dataset. We fix the number of environment divisions to 4 to study the impact of style division on performance and fix the number of style divisions to  2 to analyze the impact of environment division on performance.

Our model requires the number of style and environment divisions as input. Setting the number of style divisions to 2 achieves optimal performance by avoiding cyclic misclassifications in $\ell_{\operatorname{adv}}(\theta_f,\theta_s)$, ensuring style distribution alignment. This setting also demonstrates good alignment properties. Varying the environment division settings for different datasets accounts for varying spurious features, adapting the settings accordingly. BIRM's penalty term is more sensitive to environment divisions than IRMv1, capturing more information for improved results.

\subsection{Ablation Study}

\begin{table}[htb]
\resizebox{\columnwidth}{!}{
\begin{tabular}{lcccc|c}
\hline
\textbf{PACS}   &\textbf{Art} & \textbf{Cartoon}     & \textbf{Photo}       & \textbf{Sketch}      & \textbf{Avg}\\
\hline
Deep All  & 76.69 & 74.72 & 94.01 & 72.65 & 79.52 \\
Ours w/o $\mathbb{D}(e,\theta_f, \theta_y)$ & 78.76 & 77.65 & 94.49 & 76.73 & 81.91 \\
Ours w/o $\operatorname{sdf}(x)$ & 80.44 & 79.13 & \textbf{95.21} & 74.14 & 82.23 \\
Ours w/o $\ell_{\operatorname{adv}}(\theta_f, \theta_s)$ & 81.50 & 77.58 & 94.41 & 76.60 & 82.52 \\
Ours w/o $\ell_{\operatorname{ent}}(\theta_f, \theta_y)$ & 81.42 & 79.34 & 94.31 & 78.30 & 83.34 \\
IRSS-BIRM(Ours)              & \textbf{82.74}     & \textbf{80.51}     & 94.73     & \textbf{79.54}     & \textbf{84.38} \\
\hline
\end{tabular}
}
\caption{Ablation experiments of the IRSS-BIRM method on the PACS dataset.}
\label{tab:ablation}
\end{table}

In this section, as shown in Table \ref{tab:ablation}, we conduct a series of ablation experiments on PACS to investigate the influence of different components of our method, IRSS-BIRM. Specifically, we analyze the impact of removing the following elements: (1) $\mathbb{D}(e,\theta_f, \theta_y)$ - studying the impact of excluding spurious features on performance, (2) $\operatorname{sdf}(x)$ - examining the influence of style domain redistribution on performance, (3) $\ell_{\operatorname{adv}}(\theta_f, \theta_s)$ - investigating the effect of style domain alignment on performance, and (4) $\ell_{\operatorname{ent}}(\theta_f, \theta_y)$ - studying the influence of classification entropy on performance.

The above ablation experiments demonstrate 
the effectiveness of each component 
and the superior overall performance of our proposed method.

\subsection{Interpretability Analysis}

\begin{figure}[htb]
\centering
\includegraphics[width=\columnwidth]{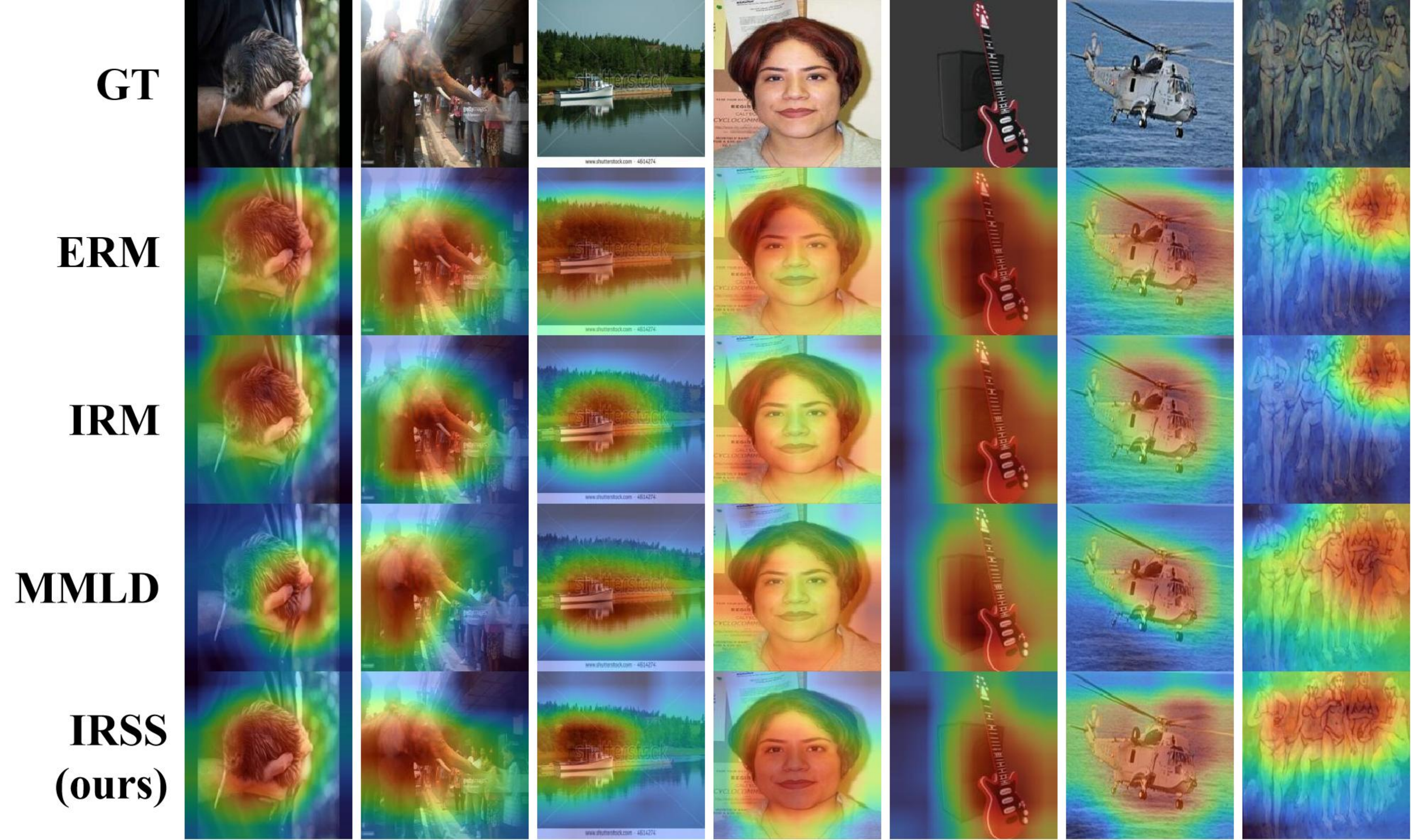} 
\caption{The original images are denoted as "ground truth" (abbreviated as GT), along with their corresponding Grad-CAM++ visualizations for the ERM, IRM, MMLD, and our \textbf{IRSS} networks.}
\label{fig:inter}
\end{figure}

We utilized Grad-CAM++ \cite{chattopadhay2018grad} for neural network interpretability and compared our method with IRM \cite{arjovsky2019invariant}, MMLD \cite{matsuura2020domain}, and ERM. Figure \ref{fig:inter} illustrates that our approach outperforms ERM, IRM, and MMLD in focusing on the target object amidst different style distributions. Furthermore, our method exhibits reduced reliance on the environment and spurious features associated with the target object, effectively eliminating their influence. This highlights the effectiveness of our approach in decoupling style features and spurious features from images.

\section{Conclusions}

This paper proposed IRSS to address the out-of-distribution (OOD) generalization problem under the setting that both style distribution shift and spurious features exist. We first propose a Structural causal model for the image generation process, which captures both style distribution shift and spurious features. The Structural causal model enables us to design our framework IRSS, which can gradually separate style distribution and spurious features from images by introducing adversarial neural networks and multi-environment optimization, thus achieving OOD generalization. Our work has achieved good experimental results on PACS, OfficeHome, and NICO datasets, demonstrating our method's effectiveness. 

\clearpage

\bigskip
\noindent

\bibliography{aaai24}

\begin{thebibliography}{32}
\providecommand{\natexlab}[1]{#1}

\bibitem[{Arjovsky et~al.(2019)Arjovsky, Bottou, Gulrajani, and
  Lopez-Paz}]{arjovsky2019invariant}
Arjovsky, M.; Bottou, L.; Gulrajani, I.; and Lopez-Paz, D. 2019.
\newblock Invariant risk minimization.
\newblock \emph{arXiv preprint arXiv:1907.02893}.

\bibitem[{Bai et~al.(2021)Bai, Sun, Hong, Zhou, Ye, Ye, Chan, and
  Li}]{bai2021decaug}
Bai, H.; Sun, R.; Hong, L.; Zhou, F.; Ye, N.; Ye, H.-J.; Chan, S.-H.~G.; and
  Li, Z. 2021.
\newblock Decaug: Out-of-distribution generalization via decomposed feature
  representation and semantic augmentation.
\newblock In \emph{Proceedings of the AAAI Conference on Artificial
  Intelligence}, volume~35, 6705--6713.

\bibitem[{Carlucci et~al.(2019)Carlucci, D'Innocente, Bucci, Caputo, and
  Tommasi}]{carlucci2019domain}
Carlucci, F.~M.; D'Innocente, A.; Bucci, S.; Caputo, B.; and Tommasi, T. 2019.
\newblock Domain generalization by solving jigsaw puzzles.
\newblock In \emph{Proceedings of the IEEE/CVF Conference on Computer Vision
  and Pattern Recognition}, 2229--2238.

\bibitem[{Chang et~al.(2020)Chang, Zhang, Yu, and
  Jaakkola}]{chang2020invariant}
Chang, S.; Zhang, Y.; Yu, M.; and Jaakkola, T. 2020.
\newblock Invariant rationalization.
\newblock In \emph{International Conference on Machine Learning}, 1448--1458.
  PMLR.

\bibitem[{Chattopadhay et~al.(2018)Chattopadhay, Sarkar, Howlader, and
  Balasubramanian}]{chattopadhay2018grad}
Chattopadhay, A.; Sarkar, A.; Howlader, P.; and Balasubramanian, V.~N. 2018.
\newblock Grad-cam++: Generalized gradient-based visual explanations for deep
  convolutional networks.
\newblock In \emph{2018 IEEE winter conference on applications of computer
  vision (WACV)}, 839--847. IEEE.

\bibitem[{Chen et~al.(2022)Chen, Zhou, Bian, Xie, Ma, Zhang, Yang, Han, and
  Cheng}]{chen2022pareto}
Chen, Y.; Zhou, K.; Bian, Y.; Xie, B.; Ma, K.; Zhang, Y.; Yang, H.; Han, B.;
  and Cheng, J. 2022.
\newblock Pareto invariant risk minimization.
\newblock \emph{arXiv preprint arXiv:2206.07766}.

\bibitem[{D’Innocente and Caputo(2019)}]{d2019domain}
D’Innocente, A.; and Caputo, B. 2019.
\newblock Domain generalization with domain-specific aggregation modules.
\newblock In \emph{Pattern Recognition: 40th German Conference, GCPR 2018,
  Stuttgart, Germany, October 9-12, 2018, Proceedings 40}, 187--198. Springer.

\bibitem[{Fan et~al.(2021)Fan, Wang, Ke, Yang, Gong, and
  Zhou}]{fan2021adversarially}
Fan, X.; Wang, Q.; Ke, J.; Yang, F.; Gong, B.; and Zhou, M. 2021.
\newblock Adversarially adaptive normalization for single domain
  generalization.
\newblock In \emph{Proceedings of the IEEE/CVF conference on Computer Vision
  and Pattern Recognition}, 8208--8217.

\bibitem[{Fang, Xu, and Rockmore(2013)}]{fang2013unbiased}
Fang, C.; Xu, Y.; and Rockmore, D.~N. 2013.
\newblock Unbiased metric learning: On the utilization of multiple datasets and
  web images for softening bias.
\newblock In \emph{Proceedings of the IEEE International Conference on Computer
  Vision}, 1657--1664.

\bibitem[{Ganin and Lempitsky(2015)}]{ganin2015unsupervised}
Ganin, Y.; and Lempitsky, V. 2015.
\newblock Unsupervised domain adaptation by backpropagation.
\newblock In \emph{International conference on machine learning}, 1180--1189.
  PMLR.

\bibitem[{Ganin et~al.(2016)Ganin, Ustinova, Ajakan, Germain, Larochelle,
  Laviolette, Marchand, and Lempitsky}]{ganin2016domain}
Ganin, Y.; Ustinova, E.; Ajakan, H.; Germain, P.; Larochelle, H.; Laviolette,
  F.; Marchand, M.; and Lempitsky, V. 2016.
\newblock Domain-adversarial training of neural networks.
\newblock \emph{The journal of machine learning research}, 17(1): 2096--2030.

\bibitem[{Gulrajani and Lopez-Paz(2020)}]{gulrajani2020search}
Gulrajani, I.; and Lopez-Paz, D. 2020.
\newblock In search of lost domain generalization.
\newblock \emph{arXiv preprint arXiv:2007.01434}.

\bibitem[{He, Shen, and Cui(2021)}]{he2021towards}
He, Y.; Shen, Z.; and Cui, P. 2021.
\newblock Towards non-iid image classification: A dataset and baselines.
\newblock \emph{Pattern Recognition}, 110: 107383.

\bibitem[{Kirillov et~al.(2023)Kirillov, Mintun, Ravi, Mao, Rolland, Gustafson,
  Xiao, Whitehead, Berg, Lo et~al.}]{kirillov2023segment}
Kirillov, A.; Mintun, E.; Ravi, N.; Mao, H.; Rolland, C.; Gustafson, L.; Xiao,
  T.; Whitehead, S.; Berg, A.~C.; Lo, W.-Y.; et~al. 2023.
\newblock Segment anything.
\newblock \emph{arXiv preprint arXiv:2304.02643}.

\bibitem[{Krueger et~al.(2021)Krueger, Caballero, Jacobsen, Zhang, Binas,
  Zhang, Le~Priol, and Courville}]{krueger2021out}
Krueger, D.; Caballero, E.; Jacobsen, J.-H.; Zhang, A.; Binas, J.; Zhang, D.;
  Le~Priol, R.; and Courville, A. 2021.
\newblock Out-of-distribution generalization via risk extrapolation (rex).
\newblock In \emph{International Conference on Machine Learning}, 5815--5826.
  PMLR.

\bibitem[{Li et~al.(2022)Li, Shen, Wang, Zhu, Li, Keutzer, and
  Zhao}]{li2022invariant}
Li, B.; Shen, Y.; Wang, Y.; Zhu, W.; Li, D.; Keutzer, K.; and Zhao, H. 2022.
\newblock Invariant information bottleneck for domain generalization.
\newblock In \emph{Proceedings of the AAAI Conference on Artificial
  Intelligence}, volume~36, 7399--7407.

\bibitem[{Li et~al.(2017{\natexlab{a}})Li, Yang, Song, and
  Hospedales}]{li2017deeper}
Li, D.; Yang, Y.; Song, Y.-Z.; and Hospedales, T.~M. 2017{\natexlab{a}}.
\newblock Deeper, broader and artier domain generalization.
\newblock In \emph{Proceedings of the IEEE international conference on computer
  vision}, 5542--5550.

\bibitem[{Li et~al.(2017{\natexlab{b}})Li, Wang, Liu, and
  Hou}]{li2017demystifying}
Li, Y.; Wang, N.; Liu, J.; and Hou, X. 2017{\natexlab{b}}.
\newblock Demystifying neural style transfer.
\newblock \emph{arXiv preprint arXiv:1701.01036}.

\bibitem[{Lin et~al.(2022)Lin, Dong, Wang, and Zhang}]{lin2022bayesian}
Lin, Y.; Dong, H.; Wang, H.; and Zhang, T. 2022.
\newblock Bayesian invariant risk minimization.
\newblock In \emph{Proceedings of the IEEE/CVF Conference on Computer Vision
  and Pattern Recognition}, 16021--16030.

\bibitem[{Lin, Lian, and Zhang(2021)}]{lin2021empirical}
Lin, Y.; Lian, Q.; and Zhang, T. 2021.
\newblock An empirical study of invariant risk minimization on deep models.
\newblock In \emph{ICML 2021 Workshop on Uncertainty and Robustness in Deep
  Learning}, volume~1, 7.

\bibitem[{Long et~al.(2015)Long, Cao, Wang, and Jordan}]{long2015learning}
Long, M.; Cao, Y.; Wang, J.; and Jordan, M. 2015.
\newblock Learning transferable features with deep adaptation networks.
\newblock In \emph{International conference on machine learning}, 97--105.
  PMLR.

\bibitem[{Mancini et~al.(2020)Mancini, Akata, Ricci, and
  Caputo}]{mancini2020towards}
Mancini, M.; Akata, Z.; Ricci, E.; and Caputo, B. 2020.
\newblock Towards recognizing unseen categories in unseen domains.
\newblock In \emph{European Conference on Computer Vision}, 466--483. Springer.

\bibitem[{Mancini et~al.(2018)Mancini, Bulo, Caputo, and
  Ricci}]{mancini2018best}
Mancini, M.; Bulo, S.~R.; Caputo, B.; and Ricci, E. 2018.
\newblock Best sources forward: domain generalization through source-specific
  nets.
\newblock In \emph{2018 25th IEEE international conference on image processing
  (ICIP)}, 1353--1357. IEEE.

\bibitem[{Matsuura and Harada(2020)}]{matsuura2020domain}
Matsuura, T.; and Harada, T. 2020.
\newblock Domain generalization using a mixture of multiple latent domains.
\newblock In \emph{Proceedings of the AAAI Conference on Artificial
  Intelligence}, volume~34, 11749--11756.

\bibitem[{Peters, B{\"u}hlmann, and Meinshausen(2016)}]{peters2016causal}
Peters, J.; B{\"u}hlmann, P.; and Meinshausen, N. 2016.
\newblock Causal inference by using invariant prediction: identification and
  confidence intervals.
\newblock \emph{Journal of the Royal Statistical Society Series B: Statistical
  Methodology}, 78(5): 947--1012.

\bibitem[{Rosenfeld, Ravikumar, and Risteski(2020)}]{rosenfeld2020risks}
Rosenfeld, E.; Ravikumar, P.; and Risteski, A. 2020.
\newblock The risks of invariant risk minimization.
\newblock \emph{arXiv preprint arXiv:2010.05761}.

\bibitem[{Torralba and Efros(2011)}]{torralba2011unbiased}
Torralba, A.; and Efros, A.~A. 2011.
\newblock Unbiased look at dataset bias.
\newblock In \emph{CVPR 2011}, 1521--1528. IEEE.

\bibitem[{Tzeng et~al.(2017)Tzeng, Hoffman, Saenko, and
  Darrell}]{tzeng2017adversarial}
Tzeng, E.; Hoffman, J.; Saenko, K.; and Darrell, T. 2017.
\newblock Adversarial discriminative domain adaptation.
\newblock In \emph{Proceedings of the IEEE conference on computer vision and
  pattern recognition}, 7167--7176.

\bibitem[{Venkateswara et~al.(2017)Venkateswara, Eusebio, Chakraborty, and
  Panchanathan}]{venkateswara2017deep}
Venkateswara, H.; Eusebio, J.; Chakraborty, S.; and Panchanathan, S. 2017.
\newblock Deep hashing network for unsupervised domain adaptation.
\newblock In \emph{Proceedings of the IEEE conference on computer vision and
  pattern recognition}, 5018--5027.

\bibitem[{Zhang et~al.(2021)Zhang, Cui, Xu, Zhou, He, and Shen}]{stablenet}
Zhang, X.; Cui, P.; Xu, R.; Zhou, L.; He, Y.; and Shen, Z. 2021.
\newblock Deep Stable Learning for Out-Of-Distribution Generalization.
\newblock arXiv:2104.07876.

\bibitem[{Zhou et~al.(2020)Zhou, Yang, Hospedales, and Xiang}]{zhou2020deep}
Zhou, K.; Yang, Y.; Hospedales, T.; and Xiang, T. 2020.
\newblock Deep domain-adversarial image generation for domain generalisation.
\newblock In \emph{Proceedings of the AAAI conference on artificial
  intelligence}, volume~34, 13025--13032.

\bibitem[{Zhou et~al.(2022)Zhou, Lin, Zhang, and Zhang}]{zhou2022sparse}
Zhou, X.; Lin, Y.; Zhang, W.; and Zhang, T. 2022.
\newblock Sparse invariant risk minimization.
\newblock In \emph{International Conference on Machine Learning}, 27222--27244.
  PMLR.

\end{thebibliography}

\clearpage

\end{document}